\documentclass[final]{cvpr}
\pdfoutput=1

\usepackage{times}
\usepackage{subfig}
\usepackage{graphicx}
\usepackage{array}
\usepackage{amsmath}
\usepackage{amssymb}
\usepackage{booktabs}
\usepackage{mathtools}
\usepackage{color}
\usepackage{pifont}
\usepackage{floatrow}
\usepackage{blindtext}
\newcommand{\cmark}{\ding{51}}
\newfloatcommand{capbtabbox}{table}[][\FBwidth]
\usepackage{hyperref}
\hypersetup{pagebackref=true,breaklinks=true,colorlinks,bookmarks=false}

\begin{document}

\title{Unsupervised Discriminative Embedding for Sub-Action Learning in Complex Activities}

\author{Sirnam Swetha$^{\star}$, Hilde Kuehne$^{\dagger}$, Yogesh S Rawat$^{\star}$, Mubarak Shah$^{\star}$ \\
$^{\star}$Center for Research in Computer Vision, University of Central Florida, Orlando, FL\\
$^{\dagger}$MIT-IBM Watson Lab, Cambridge, MA\\
}

\maketitle

\begin{abstract}


Action recognition and detection in the context of long untrimmed video sequences has seen an increased attention from the research community. 
However, annotation of complex activities is usually time consuming and challenging in practice.
Therefore, recent works started to tackle the problem of unsupervised learning of sub-actions in complex activities.
This paper proposes a novel approach for unsupervised sub-action learning in complex activities.
The proposed method maps both visual and temporal representations to a latent space where the sub-actions are learnt discriminatively in an end-to-end fashion. 
To this end, we propose to learn sub-actions as latent concepts and a novel discriminative latent concept learning {\sc (dlcl)} module aids in learning sub-actions.
The proposed {\sc dlcl} module lends on the idea of latent concepts to learn compact representations in the latent embedding space in an unsupervised way. 
The result is a set of latent vectors that can be interpreted as cluster centers in the embedding space.
The latent space itself is formed by a joint visual and temporal embedding capturing the visual similarity and temporal ordering of the data. 
Our joint learning with discriminative latent concept module is novel which eliminates the need for explicit clustering.
We validate our approach on three benchmark datasets and show that the proposed combination of visual-temporal embedding and discriminative latent concepts allow to learn robust action representations in an unsupervised setting.

\end{abstract}

\vspace{-0.6cm}
\section{Introduction}
\par Recent years have seen a great progress in video activity analysis. However, most of this research is focused on the classification of short video clips with atomic or short-range actions~\cite{ rl-a-carreira2017quo, rl-a-feichtenhofer2019slowfast, rl-a-simonyan2014two}. This is a relatively easier task when compared with analysis of untrimmed and complex video sequences
~\cite{aakur2019perceptual, Alayrac16unsupervised,  hussein2019timeception, hussein2019videograph, bf-Kuehne12, kuehne2016end, sener2018unsupervised, rl-c-shou2017cdc,  rl-c-yeung2016end}. In untrimmed video analysis, the focus is either on the problem of temporal {\em action localization} ~\cite{ad-chen2019relation, ad-long2019gaussian, ad-xu2019g, ad-zeng2019graph}, where only a set of key actions is considered in untrimmed videos; or on the task of temporal {\em action segmentation} ~\cite{kuehne2016end, sener2018unsupervised, tcfpn-ding2018weakly, as-Richard_2018_CVPR, kukleva2019unsupervised, as-Farha_2019_CVPR}, where each frame of the video is associated with a respective sub-action class as it requires to identify sub-actions and also temporally localize them.  

\begin{figure}[t]
  \includegraphics[width=2.4in, height=1.8in]{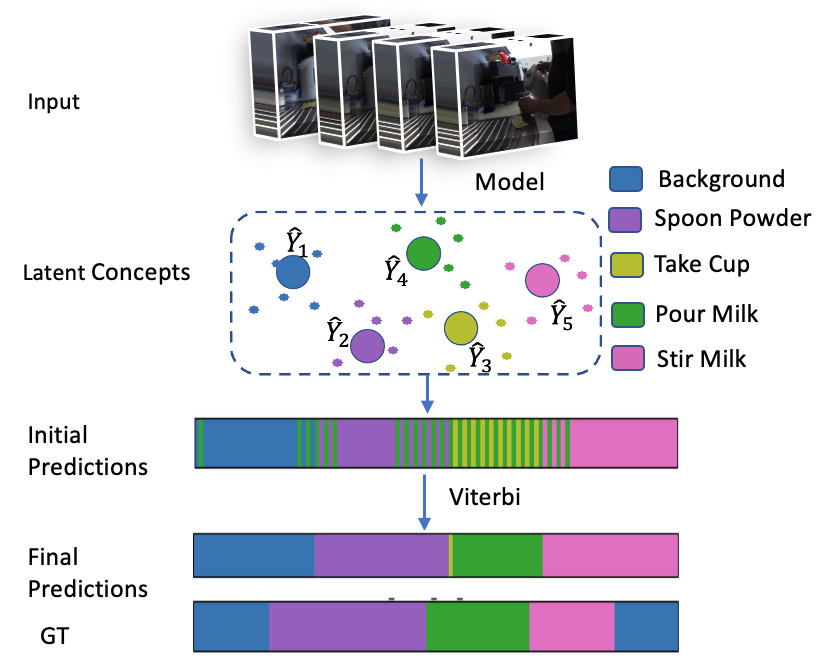}
\caption{{\small Overview of the proposed approach. Given videos of a complex activity, the proposed model learns sub-actions as latent concepts in an end-to-end manner.
The latent concept assignment for each input video segment feature forms sub-action prediction shown as `Initial Predictions', which is then refined using Viterbi to generate `Final Predictions'. Sample results for activity `Make Chocolate Milk', it can be seen that the latent concepts are able to group sub-actions. The sub-action `pour-milk' includes lifting bottle and pouring milk; the jitter can be associated to the confusion when either a chocolate/milk bottle is lifted. 
}}
\vspace{-0.3cm}
\label{fig:summ}
\end{figure}
\vspace{-0.15cm}

\par Existing works on temporal action segmentation mainly explore supervised approaches where frame-level annotations are required for all the sub-actions ~\cite{farha2019ms, bf-Kuehne12, rl-c-yeung2016end, kuehne2016end, kuehne2018hybrid, rl-c-shou2017cdc, Lea_2017_CVPR}.
However, complex activities are usually long-ranged and obtaining frame-level annotation is arduous and expensive. 
A new line of research focuses on learning these sub-actions from videos of a complex activity in an {\em unsupervised setting} ~\cite{aakur2019perceptual, Alayrac16unsupervised,  kukleva2019unsupervised, sener2018unsupervised, vidalmata2020joint} .
In the unsupervised setting, the problem is even more challenging as it requires (i) breaking down a complex activity video into semantically meaningful sub-actions; and (ii) capturing the temporal relationship between the sub-actions. 
Most approaches tackle this problem in two stages, where during the first stage an embedding based on visual and/or temporal information is learned, and in the second stage clustering is applied on this embedding space. This limits the learning ability by preventing the embedding to actually learn from clustering. At the same time, performing explicit clustering which is independent of embedding learning makes the model less efficient and does not utilize end-to-end learning.

\par To address this problem, we propose an end-to-end approach where sub-actions are learned by combining embedding and latent concepts. Here, the embedding space is trained jointly with the latent concepts leading to an effective sub-action discovery as shown in Figure~\ref{fig:summ}. To allow for such a joint training, we propose a novel discriminative latent concept learning (DLCL) module which combines latent concept learning with a contrastive loss to ensure that the sub-actions learnt in the latent embedding are distant. The resulting latent concept vectors can be interpreted as cluster centers, removing the need for explicit clustering at a later stage.

\par As the sub-actions are softly bound to the temporal position of each activity, incorporating temporal ordering is crucial. Recent works ~\cite{sener2018unsupervised, kukleva2019unsupervised} incorporated temporal embedding either by predicting the discrete temporal entities or by learning continuous temporal embedding with shallow {\sc mlp} architectures. 
In those cases, the temporal information is only given by a discrete or continuous scalar value and the joint embedding space is constructed by predicting this value from the input. 
To learn better spatio-temporal representations, we propose to use temporal position encoding ~\cite{vaswani2017attention} instead of scalar values and learn the respective embedding space by jointly reconstructing both visual and temporal representations. This embedding is further trained jointly with constrastive loss of the latent concept module, so that the embedding is also guided by and contributes to overall clustering. 

\par We evaluate our method on three benchmark datasets: Breakfast~\cite{bf-Kuehne12}, 50Salads~\cite{50s-stein2013combining} and YouTube Instructions~\cite{Alayrac16unsupervised}.
For the evaluation at test time, we follow the protocol from ~\cite{kukleva2019unsupervised} and employ the Viterbi algorithm to decode the initial sub-action predictions into coherent segments based on the ordered clustering of the sub-action latent concepts.
A detailed analysis shows the impact of the proposed elements, the reconstruction and well as the latent concept learning.

\par In summary, we propose a novel end-to-end unsupervised approach for sub-action learning in complex activities. We make the following contributions in this work:
\vspace{-0.15cm}
\begin{itemize}
    \item We propose an unsupervised end-to-end approach for sub-action learning in complex activities by jointly learning an embedding which simultaneously incorporates visual and temporal information.
    \item We learn discriminative latent concepts using contrastive loss, thus integrating clustering as part of latent embedding learning.
    \item Our method improves the state-of-the-art on three benchmark datasets.
\end{itemize}

\section{Related Work}
\par Recently, there has been a lot of interest in learning with less supervision.
This is essential for both action~\cite{rl-a-simonyan2014two, rl-a-carreira2017quo, rl-a-feichtenhofer2019slowfast} and complex activity understanding~\cite{rl-c-yeung2016end, rl-huang2016connectionist, rl-c-shou2017cdc}, as supervised approaches  require a large number of frames to be annotated in videos, which is expensive, tedious and cannot be scaled to large datasets.
Weakly supervised approaches use a video and readily available information  like accompanying text narration or audio.
Some works~\cite{rl-laptev2008learning, rl-sener2015unsupervised} use associated text narrations or scripts for learning actions in the video.
Another line of work with weak-supervision include the works where it is assumed that the  order of sub-actions is  known~\cite{tcfpn-ding2018weakly, rl-kuehne2017weakly, rl-richard2017weakly, nnvit-richard2018neuralnetwork}, however the per-frame annotations between video and sub-actions are not known during training.
Authors in ~\cite{rl-malmaud2015s} propose to use combination of audio, text and video to identify steps in instructional videos in kitchen setting. 
The performance of the above methods is highly dependent on  both the availability and quality of the text/audio alignment to video, which is not guaranteed and heavily limit their application.

\par There have been some works, in which the assumption of  weak supervision have been removed in learning of action classes. One of the first works with no supervision addressed the problem of human motion segmentation~\cite{rl-u-guerra2007language} based on sensory-motor data, and proposed an application of a parallel synchronous grammar,  to learn simple action representations similar to words in language. Later,   a Bayesian non-parametric approach to concurrently model multiple sets of time series data was proposed in ~\cite{rl-u-fox2014joint}. However, this work only focuses on motion capture data. ~\cite{rl-u-wang2015unsupervised, brattoli2017lstm}  benefit from the temporal structure of videos to fine-tune networks without any labels. Additionally, ~\cite{rl-ua-ramanathan2015learning, rl-ua-fernando2015modeling, rl-ua-cherian2017generalized, rl-ua-lee2017unsupervised} also leverage the temporal structure of videos to learn feature representation to learn actions.

\begin{figure*}[t]
  \includegraphics[width=6in, height=2.8in]{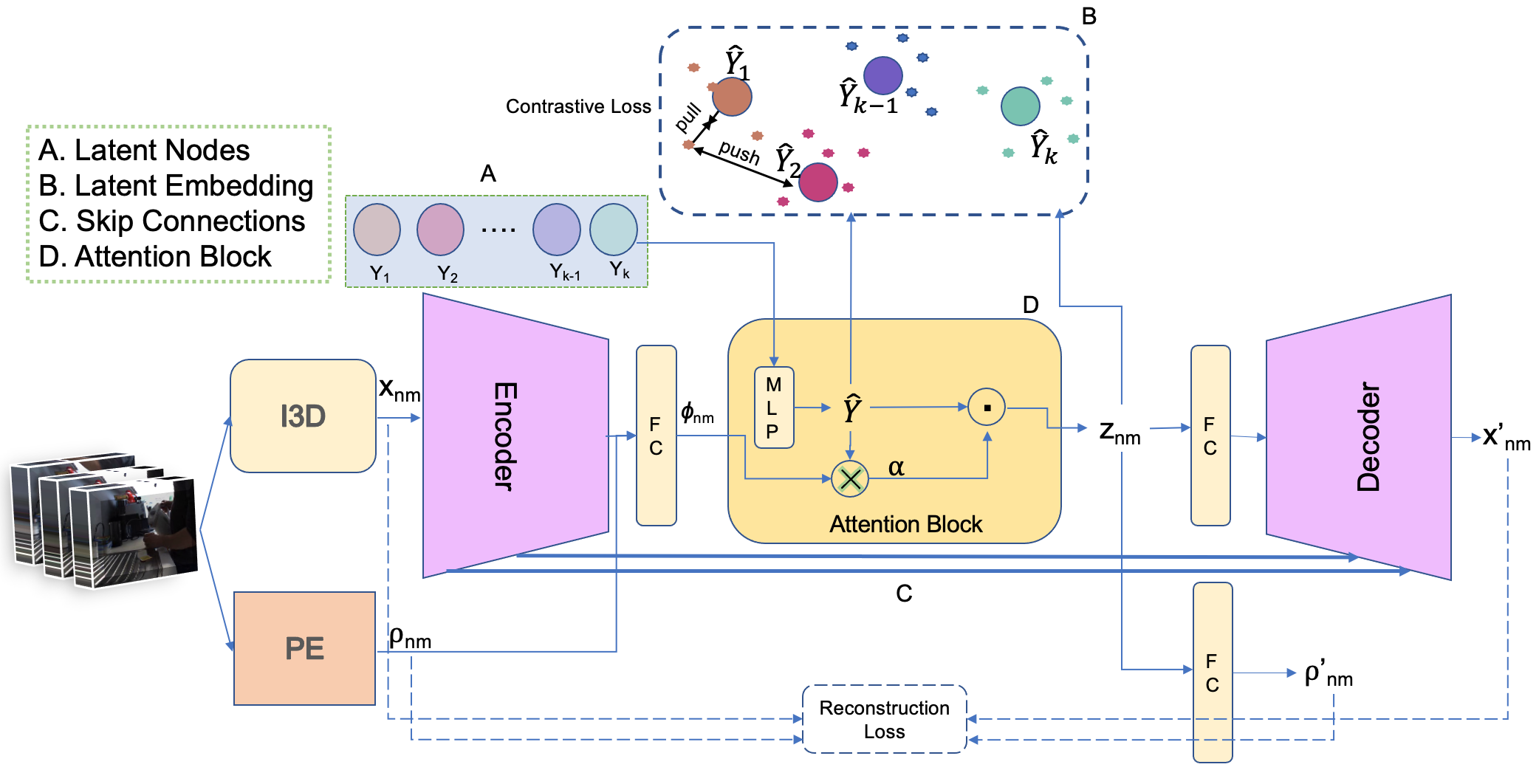}
\caption{{\small Overview of the proposed model. Given videos for a complex activity, we extract visual features $(\textsc{x}_{nm})$ and compute positional encoding vectors $(\rho_{nm})$ which are fed to the encoder to map them to a joint latent embedding for learning sub-action clusters. To learn these sub-action clusters as latent concepts $\widehat{Y}$, an attention block $(D)$ is used which takes in randomly initialized vectors $(Y)$ along with $\phi_{nm}$ and learns the latent concepts. We use contrastive loss to learn $\widehat{Y}$ discriminatively in $B$. Here $\alpha$, $z_{nm}$ and $\widehat{Y}_{k}$ represent attention, latent vector for input $\textsc{x}_{nm}$ and $k^{th}$ latent concept respectively.
}}
\vspace{-0.4cm}
\label{fig:arch}
\end{figure*}

\par Recently, unsupervised approaches have been proposed to learn sub-actions in complex activity.~\cite{sener2018unsupervised, kukleva2019unsupervised, vidalmata2020joint} propose unsupervised approaches for temporal segmentation of complex activity into sub-actions. While~\cite{aakur2019perceptual} proposes to solve a variant of the problem where the goal is to detect event boundaries, \ie event boundary segmentation for complex activities. This does not focus on identifying sub-actions, instead it learns to identify boundaries across multiple sub-actions in long videos. 
A self-supervised predictive learning framework is proposed to solve by utilizing the difference between observed and predicted frame features to determine event boundaries in complex activities.

\par In this work, we focus on solving the temporal segmentation of complex activity into sub-actions as shown in~\cite{sener2018unsupervised, kukleva2019unsupervised, vidalmata2020joint}.
In ~\cite{sener2018unsupervised}, an iterative approach is proposed that alternates between discriminative learning and generative modeling. 
For discriminative learning, they map the visual features into latent space using a `linear' transformation and compute the weight matrix which minimizes the pairwise ranking loss. For temporal ordering they use Generalized Mallows Model which models distributions over permutations as they formulate complex activity as a sequence of permutable sub-actions. 
In ~\cite{kukleva2019unsupervised}, the model incorporates the continuous temporal ordering of frames in a joint embedding space. This is achieved by training a regressor to predict the timestamp of frames in a video. The hidden layer representations are used as the embedding for clustering and the clusters are ordered with respect to their time stamp. We refer to this model as {\sc cte} (Continuous Temporal Embedding). In~\cite{vidalmata2020joint}, two-stage embedding pipeline is proposed where a next frame prediction U-Net model in stage one is combined with  with temporal discriminator in stage two followed by clustering. The temporal embedding model employed is similar to~\cite{kukleva2019unsupervised}.

\par Latent embedding learning is crucial for unsupervised learning, recently~\cite{hussein2019videograph} formulated learning graph based latent embedding using latent concepts for supervised classification of complex activities. The intuition was to model long range videos using latent concepts as graphical nodes for complex activity recognition. Inspired by their ideology of latent concept learning to model latent space, we propose DLCL as an unsupervised latent learning module with joint embedding learning to model sub-actions.

\par Most of the above works in unsupervised learning involve two stage process which does not utilize end-to-end learning making them less efficient as clustering is independent of embedding learning. In this work, we present an end-to-end model where clustering is incorporated in embedding learning using a constrastive loss. To incorporate temporal ordering we propose to use positional encodings and we also propose an effective way to unify visual and temporal representations to learn a visual-temporal embedding by jointly reconstructing visual and temporal representations.
The proposed latent embedding is not only better at capturing visual \& temporal representations but also clustering friendly. 
We demonstrate later in this paper the usefulness of the proposed model both qualitatively and quantitatively.

\section{Proposed Model}
\subsection{Overview}
\label{sec:overview}
\par Given a set of $N$ videos, $\{V_{n}\}_{n=1}^{N}$, for a complex activity, we divide each video into segments and for each segment we extract I3D features~\cite{rl-a-carreira2017quo} 
, and compute positional encoding vectors ~\cite{vaswani2017attention} as described in Section~\ref{sec:jvtl}. 
Each video 
is represented by $M_{n}$ features where $\textsc{x}_{nm}$ represent the $m^{th}$ feature in the $n^{th}$ video, and its corresponding positional encoding is represented by $\rho_{nm}$. The task is to learn the sub-actions and their ordering for each activity, i.e., by predicting sequence of a sub-action labels $l_{nm} \in \{1,2,...,K\}$ for each feature $\textsc{x}_{nm}$ for each video. The number of sub-actions labels $K$ for each activity is the maximum number of possible sub-actions as they occur in that activity.

\par Overview of our proposed model is shown in Figure~\ref{fig:arch}. First we learn an encoded representation of $\textsc{x}_{nm}$ and $\rho_{nm}$ shown as $\phi_{nm}$,
which is passed as input feature to the `Attention Block' (shown as $D$ in Figure~\ref{fig:arch}) to learn the latent concepts/clusters which are representative of sub-actions.
The attention block learns the latent concepts $\widehat{Y} \in \{\widehat{Y}_1, \widehat{Y}_2,..., \widehat{Y}_K\}$ discriminatively, where each input feature $(\phi_{nm})$ is assigned to only one latent concept.
We use a combination of reconstruction loss and constrastive loss to learn the embedding(shown as $B$ in Figure~\ref{fig:arch}). 
We believe that using a combination of both visual and temporal information in conjunction with latent concept learning to learn a latent embedding (shown as block B in fig.~\ref{fig:arch}) makes our model more robust.
We evaluate the performance of our model based on latent concept assignments for each input feature which forms `Initial Predictions'. 
Then, we model the sub-action transitions and perform Viterbi decoding to estimate optimal sub-action ordering.

\par Note that unlike previous works~\cite{kukleva2019unsupervised}, we do not perform explicit clustering, instead our model learns to cluster features in latent space as discussed in Section~\ref{sec:jvtl} \&~\ref{sec:dislc}. 
Thus eliminating the need for all the data to be available at once, our model can learn the latent concepts incrementally. 
The resulting sub-action latent concepts are temporally ordered and then each video is decoded w.r.t the above ordering given initial sub-action probability assignments for each clip to each latent concept as described in Section~\ref{sec:tmvb}.

\subsection{Joint Visual and Temporal Latent Embedding}
\label{sec:jvtl}
\par In unsupervised learning, the approach to learn clusters in latent space plays a critical role in learning semantically meaningful clusters.
We employ an encoder-decoder 
model to obtain the latent representation.
Skip-connections are included between encoder and decoder (shown as $C$ in Figure~\ref{fig:arch}), as they help to preserve commonality of an action and reduce redundant information like background in latent representation.

For incorporating temporal ordering in our model, we employ the positional encodings inspired by ~\cite{vaswani2017attention}.
We divide the video segment sequence into g equal groups and then use the ordering index to compute positional encoding vectors. Quantizing  temporal index of the video clip and using a positional encoding not only captures relative positioning but also makes it easy to generalize for highly varying video lengths. The idea of learning a mapping from features to joint visual and temporal embedding with an encoder-decoder aids in grouping clips into sub-actions in the latent space. The reconstruction loss for the auto-encoder is composed of visual features and positional encoding as shown below,
\begin{equation}
    Loss_{r} = L(\textsc{x}_{nm}, \textsc{x}'_{nm}) + \beta * L(\rho_{nm}, \rho'_{nm}), 
\end{equation}
where, $\textsc{x}_{nm}$, $\textsc{x}'_{nm}$ respectively represent input and  reconstructed visual feature; $\rho_{nm}$, $\rho'_{nm}$ respectively represent input and reconstructed positional encoding; $\beta$ is a hyperparameter and  $L$ is a loss function penalizing $\textsc{x}'_{nm}$ and $\rho'_{nm}$ for being dissimilar from $\textsc{x}_{nm}$ and $\rho_{nm}$ respectively, namely mean squared error. A combination of latent visual feature representation and the positional encodings becomes input to the `Attention Block'. In order to ensure that the learnt clusters are representative of sub-actions, the clusters have to be distant in the latent space, which is described in the next section.

\subsection{Discriminative Latent Concept Learning}
\label{sec:dislc}
\par The idea behind having this module is to learn the sub-action clusters discriminatively in the latent space in an end-to-end fashion, eliminating the need for explicit clustering. 
The attention block is inspired by ~\cite{hussein2019videograph}, which  takes an input feature $(\phi_{nm})$ and randomly initialized latent vectors $(Y)$ which is analogous to cluster center initialization as shown in Figure~\ref{fig:arch}. The latent concepts ($\widehat{Y}$) are learnt using an MLP with weight $(w)$ and bias $(b)$ i.e., it transforms the random latent vector initializations $(Y)$ to latent concepts $(\widehat{Y})$  as $\widehat{Y} = w * Y + b$. Though latent vector initialization $(Y)$ is fixed, $w$ \& $b$ are learnable parameters making the latent concepts $(\widehat{Y})$ learnable in the latent space.  These latent concepts which represent cluster centers are learned by minimizing the contrastive loss by moving features in the latent space closer to the latent concepts. The similarity  between input feature $(\phi_{nm})$ and latent concepts $(\widehat{Y})$ is measured with the dot product $\otimes$.  Then, activation function $\sigma$ is applied on the similarities to compute activation values $\alpha$ i.e., $ \alpha = \sigma(\phi_{nm} * \widehat{Y}^{T})$. Finally, the attended latent vector representation is computed as $Z_{nm} = \alpha \odot \widehat{Y}$, which captures how much each latent concept is related to the given input feature. 
However, these latent concepts tend to learn similar/overlapping concepts, which is not what we intend to learn.
Our objective in learning the latent concepts is to cluster the latent representations discriminatively. We achieve this with a contrastive loss, where the similarity between latent vectors of the same sub-action with the maximum confident latent concept is maximized, while the similarity w.r.t other latent concepts is minimized as shown in Eq~\ref{eq:cl}.
\vspace{-0.4cm}
\begin{equation}
    Loss_{d}(Z_{nm}, \widehat{Y}) = -log \frac{e^{sim(Z_{nm}, \widehat{Y}_{k^{*}})}}{\sum_{k \neq k^{*}} e^{sim(Z_{nm}, \widehat{Y}_{k})}} \label{eq:cl} 
\end{equation}
\vspace{-0.3cm}

\noindent where, $\widehat{Y}_{k}$ represents the latent concept associated with $k^{th}$ sub-action, $sim$ denotes cosine similarity and $k^{*}$ represents the latent concept with maximum confidence probability for $Z_{nm}$ as shown below
\vspace{-0.2cm}
\begin{equation}
    k^{*} = {\operatorname*{argmax}_{k}} P( k | Z_{nm})
\label{eq:sk} 
\end{equation}
\noindent where {\small $P( k | Z_{nm})=  \sigma(sim(Z_{nm}, \widehat{Y}_{k}))$} represents the confidence probability of latent vector $Z_{nm}$ for the latent concept $\widehat{Y}_{k}$, $\sigma$ is softmax activation.

\begin{figure}[t]
\includegraphics[width=3.24in, height=1.46in]{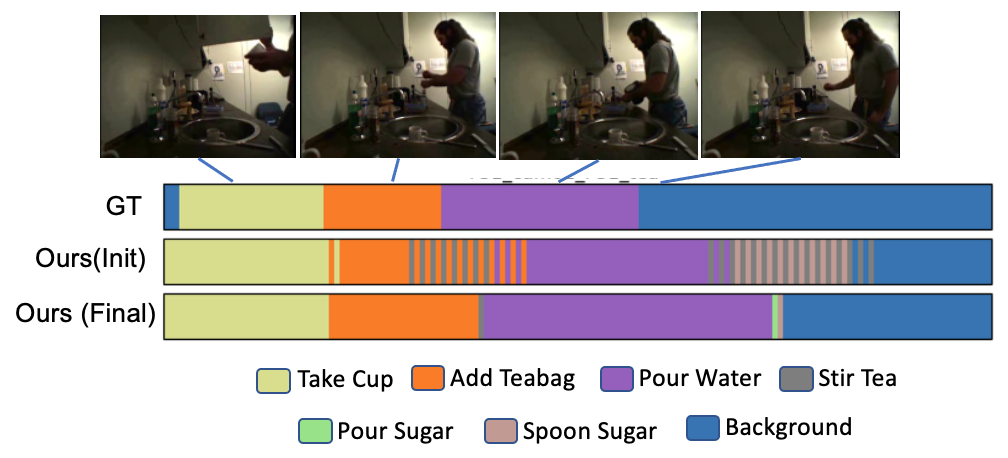}
\caption{{\small Qualitative comparison of initial predictions (w/o viterbi) and after Viterbi predictions of our approach for activity `Make Tea'. It can be seen that our model (`Init') is able to group sub-actions and also learn the ordering of sub-actions for an activity. The jitter in sub-action predictions occurs during transition from one sub-action to next, which is expected during transition. Finally, using transition modeling Viterbi decoding smoothness the jitter between sub-action transitions. 
}}
\vspace{-0.2cm}
\label{fig:qual_detail}
\end{figure}

\subsection{Overall Loss.} Total loss for learning the proposed embedding is composed of losses from Section~\ref{sec:jvtl} and~\ref{sec:dislc} as $Loss = \lambda * Loss_{r} + \gamma * Loss_{d}$

\subsection{Temporal Segmentation}

\noindent{\bf Initial Predictions}
\label{sec:ip} 
At test time, we first assign each feature in video to its respective closest latent concept vector using Eq~\ref{eq:sk}. This gives initial predictions directly based on the embedding (shown as predictions in Figure~\ref{fig:summ}). For ease of understanding, we refer to those as latent sets, analogous to clusters, from here on.

\begin{figure}[t]
  \includegraphics[width=2.6in, height=2.02in]{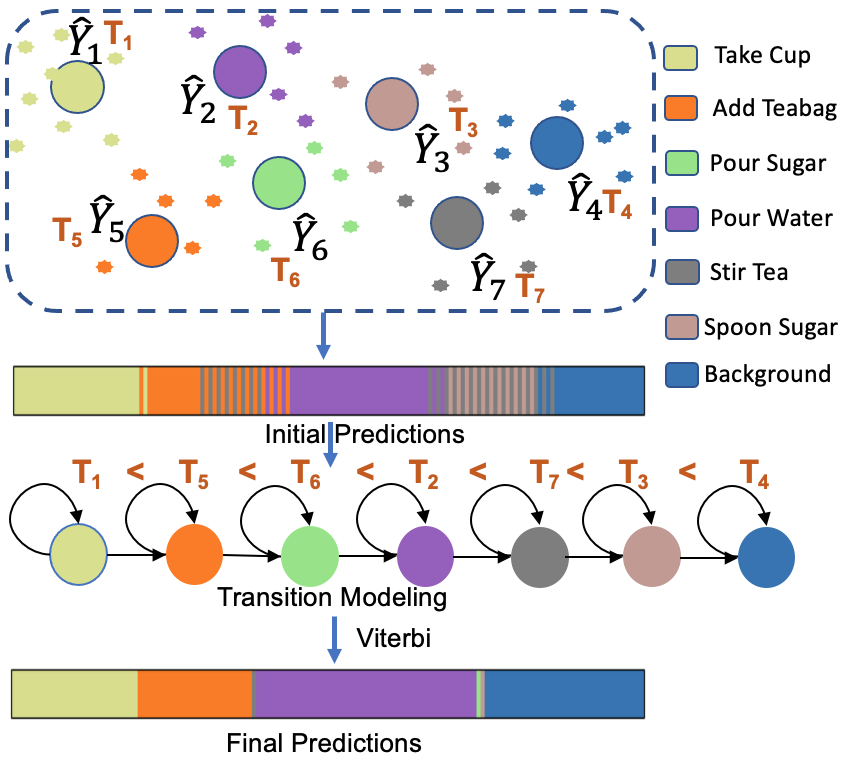}
  
\caption{{\small Brief overview of transition modeling and Viterbi decoding. Each latent concept is color coded (best viewed in color). The latent concepts are ordered w.r.t the mean time (shown as $T$) and each video is decoded into coherent segments using Viterbi algorithm based on the ordered sub-action latent concepts.
}}
\vspace{-0.3cm}
\label{fig:tmvb}
\end{figure}

\noindent{\bf Transition modeling and Viterbi decoding}\\
~\label{sec:tmvb}
\noindent Figure~\ref{fig:tmvb} represents a brief outline for transition modeling and Viterbi decoding. To allow for a temporal decoding, the global ordering of the latent sets needs to be estimated. We follow here the protocol proposed by ~\cite{kukleva2019unsupervised} and compute the mean timestamp for each set (shown as T in Figure~\ref{fig:tmvb}) and sort them in ascending order. 
The last set in the sorted ordering becomes the terminal state and using this ordering the sub-action state transition probabilities  from sub-action $i$ to $j$ are defined as $P(j|i)$ given:
\vspace{-0.3cm}
\begin{equation}
\label{eq:state-trans}
    P(j|i) = 
    \begin{dcases}
         0.5,& \text{if  $j$ immediately follows $i$ }\\
         0.5, & \text{if $i = j$} \\
         1.0, & \text{if $i = j$ \& $j$ is terminal state} \\
         0, & \text{otherwise}
    \end{dcases}
\end{equation}
\paragraph{Decoding}
\par Finally, we use the ordering and transition probabilities to compute the best path for the set ordering given the input features $\textsc{x}_{nm}$ and $\rho_{nm}$. 
Using Eq.~\ref{eq:sk} we compute the probability of each embedded input feature ($Z_{nm}$) belonging to the latent set $k$. 
We maximize the probability of the input sequence following the order defined by Eq.~\ref{eq:state-trans} to get consistent latent set assignments in a video by maximizing,
\vspace{-0.3cm}
\begin{equation}
    \label{eq:dec}
    \bar{l}_{1}^{M_{n}} = {\operatorname*{argmax}_{l_{1},...,l_{M_{n}}}}{\displaystyle \prod_{m=1}^{M_{n}} P(l_{m}|l_{m-1}) * P( l_{m} | Z_{nm})}, 
\end{equation}
\noindent where $l_{1}, ..., l_{m} \in \{1,2,...,K\}$ represent the set label sequence for $n^{th}$ video, $P( l_{m} = k | Z_{nm})$ is the probability that $Z_{nm}$ belongs to the $k^{th}$ latent set (as described in Section~\ref{sec:dislc}), $\bar{l}_{1}^{M_{n}}$ is the set label sequence for the maximum likelihood for $n^{th}$ video.

\section{Experiments}
\label{sec:exp}

For our experiments, we define a segment as a sequence of 8 frames. The video segment sequence is divided into 128 equal groups and then the ordering index is used to compute positional encoding~\cite{vaswani2017attention} for each segment. We extract {\sc i3d} features (layer `mixed\_5c') which is fed to the encoder. Our embedding dimension is 1024. We use a 3-layer encoder-decoder with Adam optimizer and the learning rate is set to $1 \times 10 ^{-4}$. We evaluate our approach on 3 datasets.

\noindent{\textbf{Breakfast Dataset}} comprises of $10$ complex activities of humans performing common kitchen activities. There are a total of $48$ sub-activities in $1,712$ videos with varying lengths based on activity and preparation style with variations in sub-action orderings.

\noindent{\textbf{50Salads Dataset}} contains videos of duration 4.5 hours for a single complex activity `making mixed salad'. 
It is a multimodal dataset, as it includes {\sc rgb} frames, depth maps and accelerometer data. However, we only use {\sc rgb} frames.
The videos in this dataset are much longer with average frame length of 10k frames and provides annotations at multiple granularity levels. 

\noindent{\textbf{YouTube Instructions Dataset}} has $5$ activities and $150$ videos with $47$ sub-actions.
These videos are taken from YouTube directly and have background segments where there is no sub-action.  The frequency and spread of background varies based on activity as well as on the person performing the task. Hence, the background segments neither have similar appearance nor have a temporal ordering. Therefore the background segments would be assigned to the latent concepts with very less confidence probability. Following protocol in ~\cite{kukleva2019unsupervised}, we consider $\tau$ percent of clips with least confidence as background. Only the foreground labeled segments along with latent concepts assignments form our initial predictions. We report results for background ratio of 60\%. 
\begin{figure}[tb]
\centering
\subfloat[Make Cereals]{\includegraphics[width=3.2in, height=0.68in]{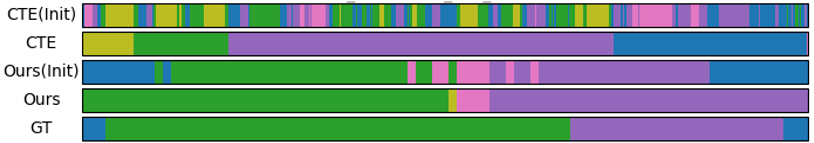}
\label{fig_first_case}
}
\hfil
\subfloat[Make Chocolate Milk]{\includegraphics[width=3.2in, height=0.68in]{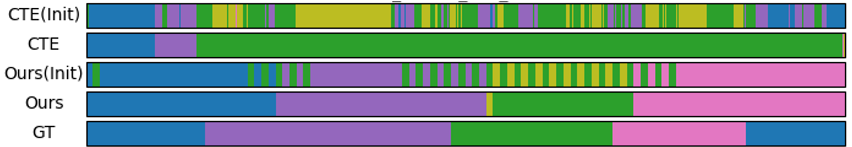}
\label{fig_first_case}
}
\caption{{\small Illustrative comparison with state-of-the-art. By comparing with {\sc cte} (Init) and Ours (Init), we show that our approach learns to model sub-actions with very few intermittent sub-action transitions leading to effective grouping of sub-actions.
Then, Viterbi decoding helps to smoothen the intermittent jitters in predictions.
We show that our method provides coherent sub-action predictions and is able to capture the orderings for sub-actions. 
}}
\vspace{-0.3cm}
\label{fig:qual_sota}
\end{figure}

\begin{table}[t]
  \caption{{\small Comparison of the proposed method to state-of-the-art on Breakfast dataset. Here, * denotes results with video-based Hungarian matching for the task event boundary segmentation.}}
  \label{tab:det_acc}
  \centering
  \small
  \begin{tabular}{lcc}
    \toprule
    {\bf Method} & {\bf F1-score} & {\bf MoF}\\
    \midrule
    &{\em  Weakly Supervised} & \\
    \midrule
    RNN-FC~\cite{rnnfc-richard2017weakly} &  & 33.3\%   \\
    TCFPN~\cite{tcfpn-ding2018weakly} &  & 38.4\%   \\
    NN-Vit~\cite{nnvit-richard2018neuralnetwork} &  & 43\%   \\
    D3TW~\cite{d3t-chang2019d3tw} &  & 45.7\%   \\
    CDFL~\cite{cdfl-li2019weakly} &  & 50.2\%   \\
    \midrule
     & {\em Unsupervised} & \\
    \midrule
    Mallow~\cite{sener2018unsupervised} & -  & 34.6\%   \\
    {\sc cte}~\cite{kukleva2019unsupervised}    & 26.4\% & 41.8\%      \\
    {\sc jvt}~\cite{vidalmata2020joint} & 29.9\% & 48.1\%      \\
    \hline
    \midrule
    {\bf Ours}     &\textbf{31.9\%}  &  47.4\%  \\
    \midrule
    \midrule
    {\sc lstm-al}~\cite{aakur2019perceptual}* & - & 42.9\%* \\
    {\bf Ours*}  & -  & \textbf{74.6\%*} \\
    \bottomrule
  \end{tabular}
\end{table}

\noindent{\textbf{Metrics}} Our model predicts a sequence of cluster labels $ \in \{1,2,...,K\}$ for each video without any correspondence 
to the $K$ ground-truth class labels. To map ground-truth and prediction label correspondences, inline with ~\cite{Alayrac16unsupervised, sener2018unsupervised, kukleva2019unsupervised}, for each activity we use the Hungarian method to find a one-to-one mapping for each cluster to exactly one sub-action and report performance after this mapping. 
In this work, we use Mean over Frames (MoF) as used by ~\cite{sener2018unsupervised,kukleva2019unsupervised} as well as F1-score used by ~\cite{Alayrac16unsupervised}. In addition,  we report Mean over class (MoC) accuracy,  as it averages the accuracy for each activity class, therefore giving equal weights to all classes irrespective of the underlying data distribution. MoF is the percentage of correct predictions computed across all activity classes together, which can be affected by the underlying activity classes distribution and biased towards dominant activity class. For F1-score, similar to previous methods, we  report the mean score over all activities. For state-of-the-art comparisons, we also evaluate our method for the task of event boundary segmentation following the protocol in ~\cite{aakur2019perceptual} and compare our method to ~\cite{aakur2019perceptual} - indicated as video-based Hungarian matching. 


\begin{table}[t]
\vspace{-0.2cm}
    \caption{{\small Comparison of MoC (Mean over class) of all activities on Breakfast dataset before and after applying Viterbi. {\sc Fv} represents Fisher Vectors.}}
    \label{tab:avg_acc}
    \centering
    \small
    \begin{tabular}{lcc}
        \toprule
        {\bf Method} & {\bf MoC} & {\bf MoC}\\
        \midrule
         & \textit{w/o Viterbi} & \textit{w Viterbi} \\
         \midrule
        {\sc cte}~\cite{kukleva2019unsupervised} with {\sc Fv} & 20.9\% & 40.1\% \\
        {\sc cte}~\cite{kukleva2019unsupervised} with {\sc i3d} & 24.8\% & 36.8\% \\
        \midrule
        \textbf{Ours with {\sc i3d}} & \textbf{37.5\%} &\textbf{46.9\%} \\
        \bottomrule
    \end{tabular}
\end{table}

\subsection{Comparison to state-of-the-art}
\par Here, we compare the proposed method to state-of-the-art approaches.
We present the accuracy comparison with recent works on Breakfast dataset in Table~\ref{tab:det_acc} and present the performance on new metric MoC in Table~\ref{tab:avg_acc}. 
Our approach achieves $47.4\%$ MoF and $31.9\%$ F1-score which is $2\%$ gain over state-of-the-art as shown in table~\ref{tab:det_acc}.
We show qualitative evaluation of the proposed approach in Figure~\ref{fig:qual_detail} \& ~\ref{fig:qual_sota}. In Figure~\ref{fig:qual_sota}, we show that our approach models the sub-actions coherently with very less intermittent sub-action transitions along with learning ordering of sub-actions for complex activity. For example, in Figure~\ref{fig:qual_detail} our model predicts `stir-tea' with intermittent transitions after `pour-water', this occurs when the person dips the tea bag in water which closely resembles to the sub-action `stir-tea' (as shown in last image in Figure~\ref{fig:qual_detail}) and then it correctly predicts background once the dip action ends (there is no annotation for `dipping tea-bag' in ground truth) indicating the goodness of the proposed sub-action learning. The intermittent transitions indicate that the model confuses to assign latent concept based on single feature and Viterbi aids in generating more coherent sub-action segments for the sequence as shown. 
Additionally, we evaluate our method for the task of event boundary segmentation and compare with the state-of-the-art approaches. Our approach out-performs the state-of-the-art MoF by a margin of $31\%$ on Breakfast dataset indicating the effectiveness of the proposed method to temporally segment meaningful sub-actions.


\begin{table}[t]
  \caption{ {\small Comparison of the proposed method to state-of-the-art unsupervised approaches on 50Salads dataset at granularity `eval'. Here, * denotes results with video-based Hungarian matching for the task event boundary segmentation.} }
  \label{tab:sal_acc}
  \centering
    \small
  \begin{tabular}{lcc}
    \toprule
    {\bf Method} & {\bf F1-score} & {\bf MoF}\\
    \midrule
    {\sc cte}~\cite{kukleva2019unsupervised}  & - & 35.5\%      \\
    {\sc jvt}~\cite{vidalmata2020joint} & - & 30.6\% \\
    \midrule
    {\bf Ours}  & \textbf{34.4\%} &  \textbf{42.2\%}  \\
    \midrule
    \midrule
    {\sc lstm-al}~\cite{aakur2019perceptual}* & - & 60.6*\% \\
    {\bf Ours*}  & - & \textbf{70.2\%*} \\
    \bottomrule
  \end{tabular}
\end{table}

\par For 50Salads dataset, we perform evaluation on granularity level `eval' and provide state-of-the-art comparison in Table~\ref{tab:sal_acc}. Our method out-performs ~\cite{kukleva2019unsupervised} by $6.67\%$ and~\cite{vidalmata2020joint} by $11.6\%$ with an F1-score of $34.37\%$. We further evaluate our method for the task of event boundary segmentation and perform state-of-the-art comparison in Table~\ref{tab:sal_acc}. We show $10\%$ gain over state-of-the-art~\cite{aakur2019perceptual} MoF, indicating our method is effective in sub-action learning for complex events. 

\par For YouTube Instructions dataset, we follow protocol in ~\cite{Alayrac16unsupervised, sener2018unsupervised, kukleva2019unsupervised} and report the performance of our approach without considering the background frames.
We achieve $42\%$ MoC \& $43.8\%$ MoF (as shown in Table~\ref{tab:acc_y}). 
This is a $4.8\%$ gain in MoF over state-of-the-art method with comparable F1-score. 
Note that ~\cite{aakur2019perceptual} reported F1-score with background frames included on YouTube Instructions Dataset. We follow the same procedure and compare our method to ~\cite{aakur2019perceptual} in Table~\ref{tab:acc_y} (indicated with *). It can be seen that our method outperforms the state-of-the-art for event boundary segmentation task showing the sub-action learning capability to identify better event boundaries.

\begin{table}[t]
  \caption{ {\small  Comparison of the proposed method to state-of-the-art unsupervised methods on YouTube Instructions dataset. Here, * denotes results with video-based Hungarian matching for the task event boundary segmentation. 
  } }
  \label{tab:acc_y}
  \centering
  \small
  \begin{tabular}{lcc}
    \toprule
    {\bf Method} & {\bf F1-score} & {\bf MoF}\\
    \midrule
    Frank-Wolfe~\cite{Alayrac16unsupervised} & 24.4\%  & 34.6\%   \\
   Mallow~\cite{sener2018unsupervised} & 27.0\%  & 27.8\%   \\
    {\sc cte}~\cite{kukleva2019unsupervised}    & 28.3\% & 39.0\%      \\
    {\sc jvt}~\cite{vidalmata2020joint} & 29.9\% & 28.2\% \\
    \midrule
    {\bf Ours} & 29.6\% &  \textbf{43.8\%}  \\
    \midrule
    \midrule
    {\sc lstm-al}~\cite{aakur2019perceptual}* & 39.7\%* & - \\
    {\bf Ours}* & \textbf{45.4\%*} & - \\
    \bottomrule
  \end{tabular}
\end{table}

\subsection{Evaluation of the Embedding.}
\label{sec:embed_eval}
To demonstrate the impact of the proposed embedding, we compare our Joint Embedding with Continuous Temporal Embedding in ~\cite{kukleva2019unsupervised} in Table~\ref{tab:avg_acc}. From Table~\ref{tab:avg_acc} (\textit{MoC w/o viterbi}), it can be seen that the proposed joint embedding outperforms the continuous temporal embedding by a huge margin of $16.6\%$. It can be seen that our `\textit{MoC w/o viterbi}' is closer to the {\sc cte} `\textit{MoC w Viterbi}' suggesting that our embedding is very effective. To emphasize that our gain in performance is due to the effectiveness of the approach and not with using {\sc i3d} features, we train ~\cite{kukleva2019unsupervised} using {\sc i3d} features by keeping the embedding dimension same as ours and compare the performance. As shown in Table~\ref{tab:avg_acc}, the MoC w/o Viterbi improves by $4\%$ by using {\sc i3d} features on {\sc cte}, while the MoC with Viterbi drops by $3\%$ with $1\%$ increase in F1-score. However, our approach still outperforms the baseline (with same embedding dimension) by huge margin indicating our approach effectiveness.

\begin{figure}[t]
\centering
  \subfloat[MoF \textit{w/o Viterbi}]{\includegraphics[width=2.68in, height=1.36in]{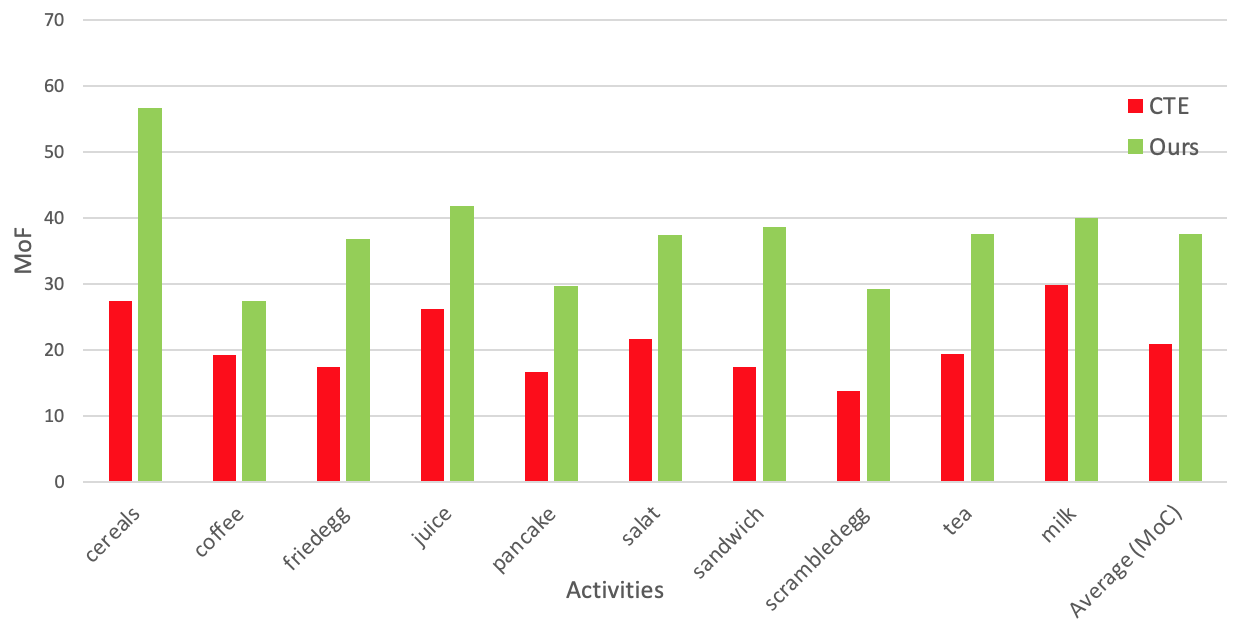}
\label{fig_first_case}
}
\hfil
\subfloat[Final MoF]{\includegraphics[width=2.68in, height=1.36in]{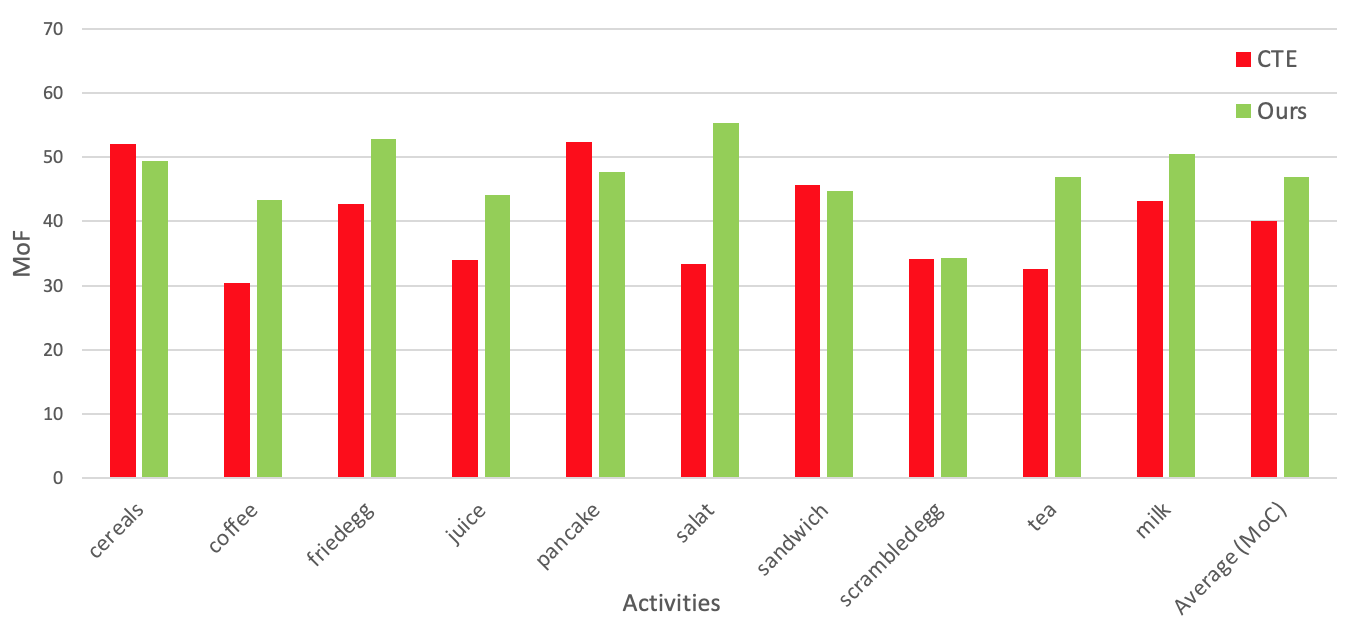}
\label{fig_second_case}
}
\caption{{\small Activity level MoF comparison on Breakfast dataset with {\sc cte}~\cite{kukleva2019unsupervised}. Last column represents the average (MoC) for all activities. (a) represents MoF for each activity without Viterbi i.e, the MoF is computed based on the learnt cluster assignments. Our method outperforms the baseline on all activities. (b) represents MoF for each activity after applying Viterbi.
}}%
\vspace{-0.3cm}
\label{fig:acc_bf}
\end{figure}

Besides dataset level comparisons, we also show activity level comparison with {\sc cte}~\cite{kukleva2019unsupervised}. Figure~\ref{fig:acc_bf} (a) shows that our joint embedding outperforms {\sc cte} on all activities indicating the significance of our joint embedding. 
We see a drop in performance for activity `making cereals' after Viterbi decoding (from Figure~\ref{fig:acc_bf}(b)), this can be attributed to the ordering of the sub-actions `take-bowl' and `pour-cereals'. For many samples in `making cereals', the sub-action `take-bowl' does not occur impacting the ordering of both sub-actions leading to drop in performance.

\subsection{Ablation Experiments}
\label{sec:ablation_exp}
\par We perform the below ablation studies on the breakfast dataset. \\
\noindent{\bf Effect of Loss Components.} To begin with, we first examine the influence of $Loss_{r}$ and $Loss_{d}$ on our model and the performances are presented in Table~\ref{tab:ablation_loss}. It can be seen that having all loss components leads to best performance. 

\noindent{\bf Effect of Discriminative learning.} The use of constrastive loss ($Loss_{d}$) helps the clusters to move apart in the latent space. This helps in obtaining more discrete boundaries in the latent space. As shown in Table~\ref{tab:ablation_loss}, the accuracy drastically reduces to $35.8\%$ ($11\% \downarrow$) indicating the importance of discriminative learning.

\begin{table}
  \caption{{\small Ablation experiments for the loss components are performed on the Breakfast dataset. $Loss_{r}$ and $Loss_{d}$ represents reconstruction loss and contrastive loss respectively. $L_{f}$ and $L_{p}$ denote the reconstruction loss for feature and positional encoding  respectively.}}
  \label{tab:ablation_loss}
  \centering
  \small
  \begin{tabular}{cc|c|c}
    \toprule
     $Loss_{r}$ & & $Loss_{d}$ & \textbf{MoC} \\
     \midrule
     $L_{f}$ & $L_{p}$ & & \\
     \midrule
     \cmark & - & - &25.7\%\\
     - & \cmark & - & 33.6\%\\
     \cmark & \cmark & - & 35.8\%\\ 
     \midrule
     \cmark & - & \cmark &  40.2\%\\
     - & \cmark & \cmark & 40.1\%\\
     \cmark & \cmark & \cmark & 46.9\%\\
    \bottomrule
    \end{tabular}
\end{table}

\noindent{\bf Effect of Positional Encoding.} Positional Encoding plays a crucial role in our model. It helps to temporally group the video clips in the latent space. As sub-actions are softly bound to the temporal position for each activity, removing reconstruction loss for positional encoding is expected to deteriorate the model performance. We observe the similar trend in Table~\ref{tab:ablation_loss}. Additionally, we perform an ablation by removing the PE component branch and train our model end-to-end. As expected, there is a significant reduction in accuracy and F1-score (as shown in Table~\ref{tab:ablation}) indicating the significance of using positional encoding.

\noindent{\bf Effect of Skip-Connections.} To assess the effectiveness of skip-connections, we report performance by removing the skip-connections and train model end-to-end. We report the performance in Table~\ref{tab:ablation}, it can be seen that w/o skip-connections, the accuracy drops considerably indicating that the skip-connections help in learning better representations.

\begin{table}[h]
 \caption{{\small Ablations experiments to evaluate the effect of PE and SC on Breakfast dataset (w/o: without, PE: Positional Encoding, SC: skip-connections).} 
 }
    \label{tab:ablation}
    \centering
    \begin{tabular}{l|cc|c}
    \toprule
     & \textbf{w/o PE} & \textbf{w/o SC} & \textbf{full} \\
     \midrule
     \textbf{MoC} & 40.9\% & 35.7\% & \textbf{46.9\%} \\
     \textbf{F1-score} & 20.3\% & 28.7\%  & \textbf{31.9\%} \\
    \bottomrule
    \end{tabular}
\end{table}

\begin{figure}
  \includegraphics[width=3.12in, height=1.46in]{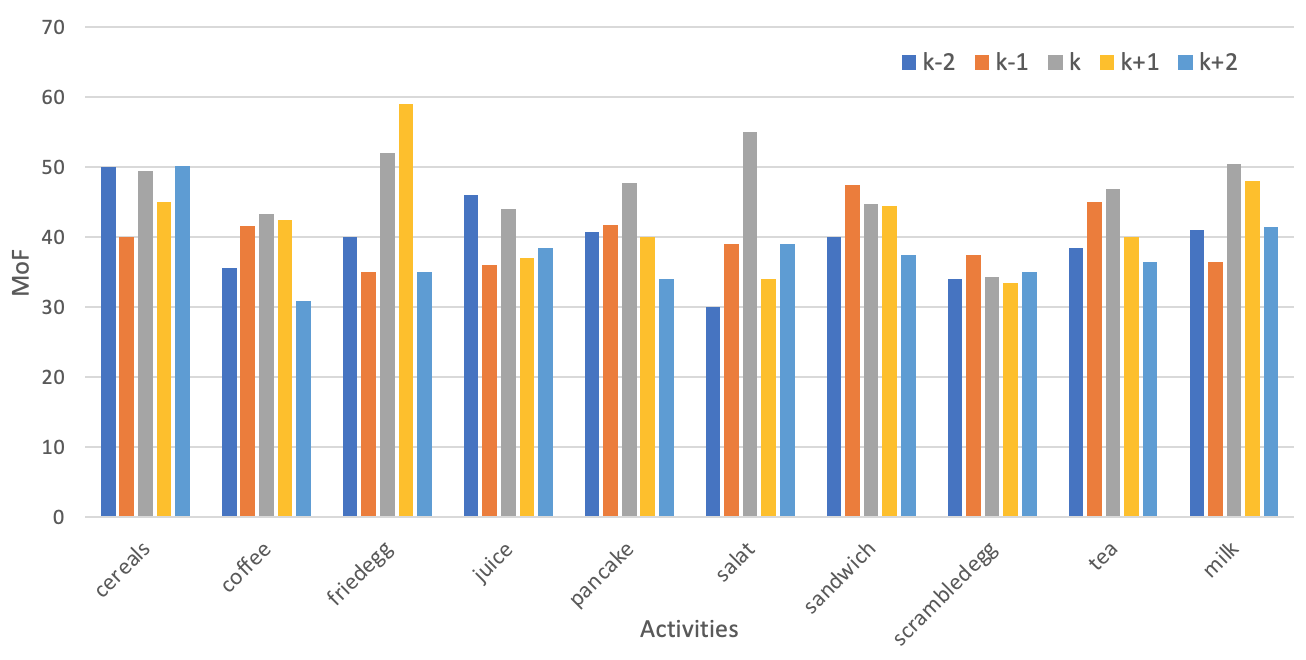}
  \caption{{\small MoF vs. \#sub-actions for all activities in Breakfast dataset. k represents the number of sub-actions from ground-truth; we vary the sub-actions for each activity and report MoF.
  }}%
 \vspace{-0.2cm}
 \label{fig:ablation_acc_vs_act}
\end{figure}

\noindent{\bf Effect of Sub-actions Cluster Size.} For all the above evaluations, the sub-action cluster size (K) is defined as mentioned in Section~\ref{sec:overview}. To analyze the impact of sub-action cluster size, we vary the number of sub-actions from $K-2$ to $K+2$ where $K$ is the number of sub-actions as per ground truth and evaluate performance. Figure~\ref{fig:ablation_acc_vs_act} shows the MoF vs number of sub-actions for each activity in Breakfast dataset. For 6 out of 10 activities we see that having K sub-actions leads to best performance.

\section{Conclusion}
In this work we proposed an end-to-end approach for unsupervised learning of sub-actions in complex activities. The main motivation behind this approach is to design a latent space to incorporate visual as well as positional encoding together. This latent space is learned via jointly training this embedding space in conjunction with a contrastive learning for clustering. We show that this allows for a robust learning that on it's own already results in a reasonable clustering of sub-actions. We then predict optimal sub-action sequence by employing the Viterbi algorithm which outperforms all the other methods. Our evaluation shows the impact of the proposed ideas and how they are able to improve the performance on this task compared to existing methods.

{\small
\bibliographystyle{unsrt}
\bibliography{cvpr}
}

\end{document}